\documentclass[conference]{IEEEtran}
\IEEEoverridecommandlockouts
\usepackage{cite}
\usepackage{amsmath,amssymb,amsfonts}
\usepackage{algorithmic}
\usepackage{graphicx}
\usepackage{textcomp}
\usepackage{xcolor}
\usepackage{algorithm}
\usepackage{multirow}
\usepackage{multicol}
\usepackage[a4paper, total={7in, 9in}]{geometry}

\def\BibTeX{{\rm B\kern-.05em{\sc i\kern-.025em b}\kern-.08em
    T\kern-.1667em\lower.7ex\hbox{E}\kern-.125emX}}
\begin{document}

\title{\vspace*{18pt}
GPTQT: Quantize Large Language Models Twice to Push the Efficiency\\
\thanks{Accepted by 11th IEEE International Conference on Cybernetics and Intelligent Systems. This work was supported by the National Natural Science Foundation.China (No.62173300)}
}

% \author{
% \IEEEauthorblockN{1\textsuperscript{st} Yipin Guo}
% \IEEEauthorblockA{\textit{College of Control Science and Engineering} \\
% \textit{Zhejiang University}\\
% Hangzhou, China \\
% guoyipin@zju.edu.cn}
% \and
% \IEEEauthorblockN{2\textsuperscript{nd} Yilin Lang}
% \IEEEauthorblockA{\textit{College of Control Science and Engineering} \\
% \textit{Zhejiang University}\\
% Hangzhou, China \\
% langyilin@zju.edu.cn}
% \and
% \IEEEauthorblockN{3\textsuperscript{rd} Qinyuan Ren}
% \IEEEauthorblockA{\textit{College of Control Science and Engineering} \\
% \textit{Zhejiang University}\\
% Hangzhou, China \\
% renqinyuan@zju.edu.cn}
% }

\author{Yipin Guo, Yilin Lang, Qinyuan Ren\\
\textit{College of Control Science and Engineering, Zhejiang University}, Hangzhou, China\\
\tt\small \{guoyipin, langyilin, renqinyuan\}@zju.edu.cn}

\maketitle

\begin{abstract}
Due to their large size, generative Large Language Models (LLMs) require significant computing and storage resources. This paper introduces a new post-training quantization method, GPTQT, to reduce memory usage and enhance processing speed by expressing the weight of LLM in 3bit/2bit. Practice has shown that minimizing the quantization error of weights is ineffective, leading to overfitting. Therefore, GPTQT employs a progressive two-step approach: initially quantizing weights using Linear quantization to a relatively high bit, followed by converting obtained int weight to lower bit binary coding. A re-explore strategy is proposed to optimize initial scaling factor. During inference, these steps are merged into pure binary coding, enabling efficient computation. Testing across various models and datasets confirms GPTQT's effectiveness. Compared to the strong 3-bit quantization baseline, GPTQT further reduces perplexity by 4.01 on opt-66B and increases speed by 1.24$\times$ on opt-30b. The results on Llama2 show that GPTQT is currently the best binary coding quantization method for such kind of LLMs.
\end{abstract}


\section{Introduction}

Pre-trained large language models (LLMs), especially from the Transformer family like the Generative Pre-trained Transformer (GPT)\cite{yenduri2023gpt} and Open Pre-trained Transformer (OPT)\cite{zhang2022opt}, excel in complex language tasks. Their success has generated considerable academic and practical interest.

However, their high computational and storage demands pose a significant adoption barrier. For instance, GPT-3 175B's parameters, even when compressed to float16, require about 326GB for inference. This exceeds the capacity of advanced single GPU units, often necessitating expensive multi-GPU configurations.

Quantization technology reduces model parameter and activation precision, e.g., substituting float32 with int8 cuts storage needs by 4$\times$ \cite{Xiao2023SmoothQuant,Dettmers2022int8()}. While applying low-bit weight quantization to LLMs promises memory savings, it introduces significant challenges. Quantization-aware training (QAT) \cite{Dettmers2023qlora,xu2024qalora}, though effective, is impractical for LLMs due to high training costs. Conversely, post-training quantization (PTQ) suffers from substantial accuracy losses in low-bit configurations, compromising model performance.

% \begin{figure}[h]
%     \centering
%     \includegraphics[width=0.5\linewidth]{figure/cal_memory_boundary.png}
%     \vspace{-1.0em}
%     \caption{Computational and communication constraints differ based on the level of concurrency\cite{Song2023mit65940}. when bs=1, the operation primarily faces limitations due to communication bandwidth. Under such conditions, opting to quantize only the weights proves to be an effective strategy in reducing the volume of data that needs to be communicated.}
%     \label{fig:boundary}
%     \vspace{-1.5em}
% \end{figure}

Due to communication bandwidth constraints, keeping activations in float16 while quantizing weights to low bits has proven effective in local settings with low concurrency. AWQ\cite{lin2023awq} keeps 1\% of critical weights in float16, quantizing the rest in a w4a16 LLM configuration. GPTQ\cite{frantar2023optq} uses second-order information for error compensation, resulting in acceptable accuracy losses under w3a16. BiLLM\cite{huang2024billm} integrates binary weights into LLMs, achieving 1.11bits with unstructured sparsity. However, GPTQ's linear quantization and maintenance of float16 activations lack effectiveness. Furthermore, BiLLM’s aggressive quantization complicates achieving tangible hardware performance gains. BCQ\cite{kwon2021bcq} first applies binary coding\cite{rastegari2016xnornet} to LLMs but only iteratively optimizes quantized MSE weight error, leading to significant accuracy losses.

GPTQT aims to improve the accuracy of post-training quantized LLMs and achieve significant performance enhancements on general-purpose GPUs. It introduces a new binary-coding method, offering enhanced representational capacity within the same bit allocation.

Specifically, GPTQT quantizes LLMs in two stages. First, it linearly quantizes weights to a higher bit resolution, then converts these to lower bit binary-coded weights. This change in representation range compels GPTQT to reassess and adjust the scaling factor to maintain and restore accuracy. In inference, these two steps are merged into pure binary coding, which can use efficient computing methods. Our contributions are summarized as follows:

\begin{itemize}
    \item Introduce GPTQT, a novel post-training quantization approach that converts LLM weights to low-bit binary coding via a heterogeneous, progressive, two-stage quantization process.
    
    \item Propose a new re-exploration scheme for the scaling factor in response to changes in representation range during progressive quantization, optimizing accuracy.
    
    \item Show that weights processed by GPTQT eliminate intermediate states in inference, enabling the use of an efficient binary-coding weight calculation method to significantly boost processing speed.
    
\end{itemize}

 \section{GPTQT: quantize LLM twice}

    \begin{figure*}[h]
        \centering
        \includegraphics[width=0.9\hsize]{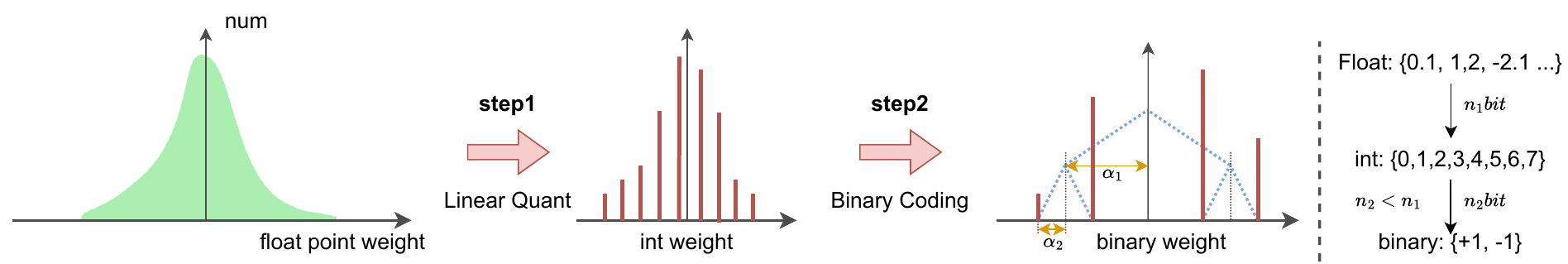}
        \vspace{-0.5em}
        \caption{GPTQT: Quantize Weight Twice. Initially, the fp16 weight model is quantized to a relatively high bit number (3 bits shown) using linear quantization. Subsequently, the resulting int-type weight is further reduced to fewer bits (2 bits depicted) using binary coding.}
        \label{fig:main}
        \vspace{-0.5em}
    \end{figure*}

\subsection{Background}

    \textbf{GPTQ.} Based on Optimal Brain Quanization\cite{frantar2023OBC}, GPTQ perform error compensation with Hessian Matrix $H_F = 2X_FX_F^T$, where $F$ denotes the set of remaining full-precision weights. It first confirms the quantization parameters row-wisely, then quantifies the weights column by column, and updates the unquantized weights in real time. 
    \begin{equation}
        \label{equ:gptq_aim}
         Goal: argmin_{w_q} \frac{(quant(w_q)-w_q)^2}{H_F^{-1}} ,
    \end{equation}
    \begin{equation}
        \label{equ:gptq_compensate}
         \delta_F = -(w_q-quant(w_q))([H_F^{-1}]_{qq})^{-1}(H_F^{-1})_{:,q}.
    \end{equation}
    The optimization goal is Equ.\ref{equ:gptq_aim}, and the compensation is Equ.\ref{equ:gptq_compensate} . The subscript $q$ represents the weight being quantized.

    \textbf{Binary Coding Weight}. Unlike linear quantization, which evenly distributes several points across the original data range, binary coding weight sets unequal representation intervals. Specifically, it represents the weight as several binarized bits $b_i(\in \{-1,+1\})$, with each bit corresponding to a floating point number  $\alpha_i$. Its quantization and dequantization process is shown as Equ. \ref{equ:bcw}.
    \begin{equation}
        \label{equ:bcw}
        \begin{matrix}
        quant: b_i=sign(r_{i-1}), \alpha _i=\frac{r_{i-1}^Tb_i}{n}, r_i=w-\sum_{j=1}^{i-1}  \alpha _jb_j, \\
        dequant: w_{dq} = \sum_{i=1}^{n}  \alpha _ib_i.
        \end{matrix}
    \end{equation}

    Where $n$ is the number of quantization bits, $w_{dq}$ is the inverse quantization result, and $r$ is the current quantization residual.

    According to Equ.\ref{equ:bcq}, BCQ\cite{kwon2021bcq} uses an iterative method to alternately optimize $\alpha$ and $b$, which reduces the MSE error, but still causes a non-negligible loss of accuracy on LLM. It will be compared as a baseline.
    \begin{equation}
        \label{equ:bcq}
        [\alpha _1,...\alpha_n]=((B_n^TB_n)^{-1}B_n^Tw)^T.
    \end{equation}

\subsection{Quantize Weight Twice}

    Intuitively, binary coding methods like BCQ, optimized from scratch, should outperform Linear Quantization. Yet, experiments reveal that integrating BCQ directly into the GPTQ method fails. GPTQ, which initially uses linear quantization, sets quantization parameters, then quantizes column by column, compensating earlier errors in subsequent columns. In this mechanism, weights considered for quantization parameters differ from those actually quantized. Therefore, the BCQ method that overfits the original weights loses its generalization to GPTQ at this situation.

    To address this, GPTQT employs a two-step progressive quantization as shown in Fig. \ref{fig:main}. Initially, linear quantization is used to quantize the weights to a relatively high bits as $W_{int}$. Subsequently, critical points from this output are selected and re-encoded using low-bit binary coding.

    \vspace{-0.5em}
    \begin{equation}
        \label{equ:gptqt}
        \begin{matrix}
            step1: W_{int}=round(\frac{W}{S}-qbias),  \\ \\
            step2: argmin_{W_q}(W_qX), W_q= BinrayCoding(W_{int}).
        \end{matrix}
    \end{equation}
    \vspace{-0.5em}

    In this process, $S$ represents the scaling factor for linear quantization, $qbias$ is the quantization offset, $W_{int}$ denotes the intermediate high-bit value from the first quantization step, and $Wq$ is the final quantized weight.

    To identify critical segments of $W_{int}$, GPTQT employs a grid search to minimize output errors, serving as the optimization criterion.

    The first step maintains finer representable distances and the generalizability of linear quantization. It completes quantization with 5 bits. The second step reduces $W_{int}$ to a lower bit, for example, to 3 bits.

    During the second step, from all integers representable after the first step, select a few and convert the remainder to the nearest selected number. For instance, if the weight is quantized to 3 bits initially, the numbers ${0,1,2,3,4,5,6,7}$ can be represented. For the binary coding step, you might select numbers representable by 2 bits, such as ${0,1,6,7}$."
    
    \vspace{-0.5em}
    \begin{equation}
        \label{equ:round}
            \begin{matrix}
            W_{int}=[0,2,3,1,1,6,5] \in \{0\sim 2^{n-1}\}
            \\
            BCchoice = [0,1,6,7]
            \\ \\
            W_q = BinaryCoding(W_{int}, BCchoice)
            \\
            =[0,1,1,1,1,6,6]
            \end{matrix}
    \end{equation}
    \vspace{-0.5em}

    This process involves rounding from one integer to the nearest target integer, denoted as $BCchoice$, which represents the target expressible in binary coding. Since the first step initially reduces the number of weight bits, the feasible $BCchoice$ options are limited, allowing for a sequential trial of each possibility.

    \subsection{Re-explore Scale Factor}

    \begin{figure}[h]
        \centering
        \includegraphics[width=\linewidth]{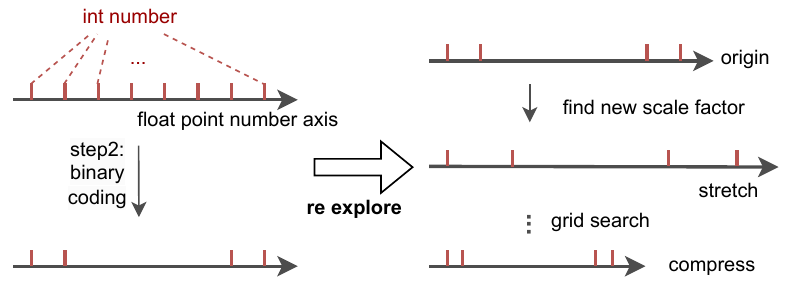}
        \vspace{-2.0em}
        \caption{Re-exploring scale factor.}
        \label{fig:reexplore}
        \vspace{-0.5em}
    \end{figure}

    In the initial step, floating point weight information is reduced to a constrained set of integer numbers, making each one critical. Yet, the second step further simplifies this information, instead of extracting directly from the original data, leading to additional information loss.

    To address the issues of overfitting or significant information loss, GPTQT re-explore the scaling factor used in the first step.

    We contend that the scaling factor $S$, determined in step 1, becomes inadequate due to the modifications in the second step of quantization. This second step introduces gaps in the uniformly distributed integer axis. Some gaps may be too wide, significantly increasing quantization errors.

    To mitigate this, we adjust the numerical axis like a spring, allowing it to stretch and compress to an optimal degree, as shown in Fig.\ref{fig:reexplore}, to identify the best scale factor as per Equ.\ref{equ:gptqt}.

    For instance, if the weight is quantized to $n$ bits in step 1, then $S=(W_{max}-W_{min}) / (2^n-1)$. The exploration is conducted within the range corresponding to $n-1$ and $n+1$ bits:

    \begin{equation}
        \label{equ:reexplore}
        \hat{S}\in (\frac{W_{max}-W_{min}}{2^{n+1}-1},\frac{W_{max}-W_{min}}{2^{n-1}-1}) 
    \end{equation}

    Where $\hat{S}$ represents the scaling factor determined through re-exploration, with $W_{max}$ and $W_{min}$ being the maximum and minimum values of the original weights, respectively.

    GPTQT conducts a grid search within the range specified in Equ.\ref{equ:reexplore} to identify the optimal value.

    \subsection{Intermediate Steps Can be Fused in Inference}

    \begin{figure}[h]
        \centering
        \includegraphics[width=\linewidth]{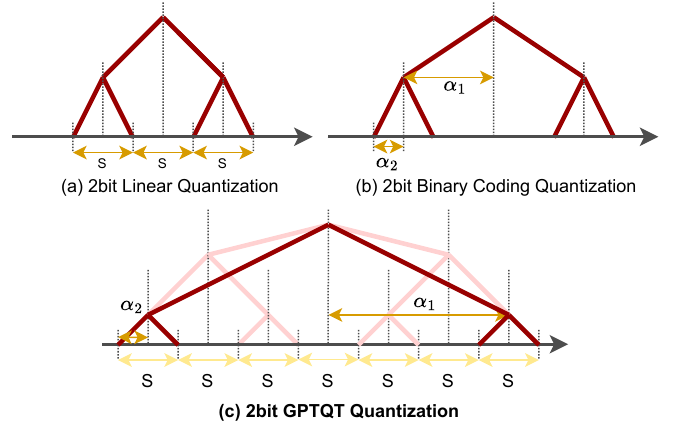}
        \vspace{-2.0em}
        \caption{Binary coding is a unique variant of linear quantization, structured in a tree-like form. GPTQT selects specific nodes and cotyledons from the linear quantization tree to create a new binary coding tree, thus bypassing intermediate steps during inference. \textit{Dark colors} indicate final results while \textit{light colors} denote intermediate results.}
        \label{fig:fuse}
        \vspace{-0.5em}
    \end{figure}

    Binary coding pushes the gaps between the quantized weight value in continuous space to be different. Therefore, Linear quantization can be regarded as a special form of binary coding. Specifically, consider wanting to quantize weights to 2 bits using Linear quantization; the available integer weights would be $W_{int} \in [-2,-1,0,1]$. To convert these back to floating point numbers, a scaling factor $S$ and a quantization bias $qbias$ are employed. The resulting dequantized weight is then expressed as $\hat{W} = S \times W_{int} + qbias$. If set:

    \vspace{-0.5em}
    \begin{equation}
        \label{equ:linear_eql_bcq}
        \alpha _2 = 2^{-1}S,\space  \alpha _1 = 2^{0}S,
    \end{equation}
    % \vspace{-1.0em}
    the points representable by binary coding are equivalent to those by linear quantization. Given this characteristic, the entire quantization process can be consolidated into a single binary coding expression during inference.

    For instance, in Fig.\ref{fig:fuse} (c), the process initially completes linear quantization with 3 bits, followed by 2-bit binary coding. During fusion, the result of linear quantization is first transformed into a binary coding representation, and subsequently, the scaling factor $S$ and partial bit $\alpha$ are integrated.

    \begin{equation}
        \label{equ:fuse_step1}
        \begin{matrix}
        \textbf{step1}:W_{int}\in \{0,1,2,3,4,5,6,7\}(3bit),
         \\
        W_{int}=\sum_{i=1}^{3} \alpha_ib_i+3.5,
        \\
        \alpha_i=2^{i-2}, b_i=\{+1,-1\},
        \end{matrix}
    \end{equation}

    \begin{equation}
        \label{equ:fuse_step2}
        \begin{matrix}
        \textbf{step2}:W_q=\sum_{i=1}^{2}\hat{\alpha_i}\hat{b_i} + 3.5,
        \\
        \hat{\alpha_1}=2^{-1}, \hat{\alpha_2}=2^{-0}+2^{1}, \hat{b_i}\in\{+1,-1\}.
        \end{matrix}
    \end{equation}

    Where $\hat{\alpha_i},\hat{b_i}$ is the fused width floating point number and binary value.

    Then we also take the dequantization process of linear quantization into consideration, and the final fused binary coding is expressed as:
    \begin{equation}
        \label{equ:fuse_step3}
        \begin{matrix}
        \textbf{fused}:W_q=\sum_{i=1}^{2}\hat{\alpha_i}\hat{b_i} + (3.5S+qbias),
        \\
        \hat{\alpha_1}=2^{-1}S, \hat{\alpha_2}=(2^{-0}+2^{1})S, \hat{b_i}\in\{+1,-1\}.
        \end{matrix}
    \end{equation}

    This approach enables the use of efficient binary coding calculation methods like LUT-GEMM\cite{park2024lutgemm} during inference.

    Given the limited binary coding options available in step 2, directly performing a grid search to optimize Equ.\ref{equ:gptqt} is feasible when the bit number of $Wq$ does not exceed 4 bits.

    % In summary, the quantization process of GPTQT is detailed in Alg. \ref{alg:gptqt}.

    % \begin{algorithm}[h]
    %    \caption{GPTQT}
    %    \label{alg:gptqt}
    %     \begin{algorithmic}
    %        \STATE {\bfseries Input:} Weight $W$, calibration data $X$
    %        \STATE $\hat{S}=None, BestBinaryCoding = None$
    %        \STATE $MinError=inf, qbias = W_{min}$
    %        \REPEAT 
    %        \FOR{$S'$ in $linspace(\frac{W_{max}-W_{min}}{2^{n+1}-1},\frac{W_{max}-W_{min}}{2^{n-1}-1})$}
    %        \FOR{$BCchoice$ in $BinaryCodingAllChoice(W_{int})$}
           
    %             \STATE $W_{int} = LinearQuant(W,S',qbias)$
    %             \STATE $W_q' = BinaryCoding(W_{int}, BCchoice)$
    %             \IF{$error(W_q'X, WX) < MinError$}
    %                 \STATE $\hat{S} = S', BestBinaryCoding=BCchoice$
    %             \ENDIF                
    %        \ENDFOR
    %        \ENDFOR
    %        \STATE $W_q = GPTQ(W, \hat{S}, BCchoice)$
    %        \STATE Fuse quantization param. 
    %        \UNTIL {All layers are quantized.}
    %     \end{algorithmic}
    % \end{algorithm}
\section{Experiment}

    \begin{table*}[t]
  \centering
  \caption{OPT perplexity (the lower the better) results on WikiText2.}
  \vspace{-1em}
  \renewcommand{\arraystretch}{1.15} % increase vertical spacing
  \resizebox{0.75\linewidth}{!}{
    \begin{tabular}{c|c|cccccccc}
    \hline
    \hline
    \textbf{Method} & \textbf{Bits} & \textbf{125M} & \textbf{350M} & \textbf{1.3B} & \textbf{2.7B} & \textbf{6.7B} & \textbf{13B} & \textbf{30B} & \textbf{66B}\\
    \hline
    \hline
    full & 16 & 27.65 & 22.00 & 14.63 & 12.47 & 10.86 & 10.13 & 9.56 & 9.34 \\
    \hline

    RTN & \multirow{4}[2]{*}{3} & 1.3e3 & 64.57 & 1.3e4 & 1.6e4 & 5.8e3 & 3.4e3 & 1.6e3 & 6.1e3 \\
    BCQ &  & 60.00 & 42.32 & 49.09 & 17.55 & 20.027 & 12.50 & 139.9 & 100.33 \\
    GPTQ & & 53.85 & 33.79 & 20.97 & 16.88 & 14.86 & 11.61 & 10.27 & 14.16 \\
    \textbf{GPTQT} & & \textbf{90.31} & \textbf{27.41} & \textbf{17.23} & \textbf{14.88} & \textbf{14.23} & \textbf{12.98} & \textbf{10.10} & \textbf{10.15} \\
    \hline

    RTN & \multirow{4}[2]{*}{2} & 5.4e3 & 2.8e4 & 1.1e4 & 9.5e3 & 2.8e4 & 1.9e5 & 1.6e5 & 1.7e5 \\
    BCQ &  & 484.32 & 2.0e3 & 3.8e3 & 616.3 & 1.7e4 & 4.9e3 & 7.7e3 & 6.2e3 \\
    GPTQ & & 2.4e3 & 1.0e4 & 4.7e3 & 6.3e3 & 442.6 & 126.09 & 71.70 & 20.91 \\
    \textbf{GPTQT} & & \textbf{5.6e3} & \textbf{706.3} & \textbf{248.3} & \textbf{56.03} & \textbf{91.21} & \textbf{27.82} & \textbf{13.26} & \textbf{12.53} \\

    \hline
    \hline
    \end{tabular}%
  }
  \vspace{-1em}
  \label{tab:main}%
\end{table*}%

    \begin{table*}[t]
  \centering
  \caption{Llama2 and Bloom perplexity results on WikiText2.}
  \vspace{-1em}
  \renewcommand{\arraystretch}{1.15} % increase vertical spacing
  \resizebox{0.9\linewidth}{!}{
    \begin{tabular}{c|c|cc|ccccc}
    \hline
    \hline
    \textbf{Method} & \textbf{Bits} & \textbf{Llama2-7b} & \textbf{Llama2-13b} & \textbf{Bloom-560m} & \textbf{Bloom-1.1b} & \textbf{Bloom-1.7b} & \textbf{Bloom-3b} & \textbf{Bloom-7.1b} \\
    \hline
    \hline
    full & 16 & 5.47 & 4.88 & 22.42 & 17.69 & 15.39 & 13.48 & 11.37  \\
    \hline

    BCQ &  \multirow{3}[1]{*}{3} & 117.41 & 6872.46 & 48.87 & 46.37 & 24.723 & 21.02 & 17.29   \\
    GPTQ & & 32.31 & 25.08 & 32.31 & 25.08 & 21.11 & 17.40 & 13.47   \\
    \textbf{GPTQT} & & \textbf{12.80} & \textbf{6.27} & \textbf{38.39} & \textbf{22.56} & \textbf{20.97} & \textbf{17.53} & \textbf{12.41} \\

    \hline
    \hline
    \end{tabular}%
  }
  \vspace{-1em}
  \label{tab:otherModel}%
\end{table*}%

The experiment primarily assesses the language generation task using \textbf{Perplexity} as the performance metric for language models, which are notably sensitive to model quantization\cite{Yao2022ZeroQuant}. Perplexity evaluates how effectively a probability distribution or language model predicts a sample, essentially measuring the model's surprise when predicting the next word in a sequence. \textit{Lower perplexity values signify superior performance}, indicating greater model confidence and less surprise with the data encountered. Additionally, GPTQT not only reduces storage requirements but also enhances processing speed on general-purpose GPUs. To demonstrate this, the time taken to generate a single token is recorded.

\subsection{Setup}

    GPTQT is implemented in Pytorch and integrates with HuggingFace’s OPT, BLOOM\cite{workshop2023bloom}, and Llama\cite{touvron2023llama} model families. Models with fewer than 13B parameters are quantized on an A5000 GPU, while larger models are processed on an NVIDIA A100 GPU with 80GB due to memory constraints. Calibration employs 128 random slices of 2048 tokens each from the datasets.

    Testing primarily utilized the WikiText2\cite{merity2016wikitext} and PTB\cite{dinarelli2019ptb} datasets.
    
    Token generation speed was evaluated on an A5000 GPU. The test method involved generating a sequence of 128 tokens with a batch size of 1 and timing this to calculate the average token generation time. Notably, GPTQT also reduces weight to 3 bits when measuring speed, aligning the communication overhead with GPTQ. The speed improvement solely results from the adoption of more efficient binary coding calculation methods. Given that the fp16 model consumes more storage, model parallel technology was employed with as few GPUs as possible.

\subsection{Result on OPT}
    \begin{table*}[t]
  \centering
  \caption{OPT perplexity results on PTB.}
  \vspace{-1em}
  \renewcommand{\arraystretch}{1.15} % increase vertical spacing
  \resizebox{0.65\linewidth}{!}{
    \begin{tabular}{c|c|ccccccc}
    \hline
    \hline
    \textbf{Method} & \textbf{Bits} & \textbf{125M} & \textbf{350M} & \textbf{1.3B} & \textbf{2.7B} & \textbf{6.7B} & \textbf{13B} & \textbf{30B} \\
    \hline
    \hline
    full & 16 & 32.55 & 26.08 & 16.96 & 15.11 & 13.08 & 12.34 & 11.84 \\
    \hline

    BCQ &  \multirow{3}[1]{*}{3} & 70.59 & 52.53 & 112.06 & 22.98 & 23.61 & 14.74 & 258.74  \\
    GPTQ & & 67.28 & 40.10 & 26.69 & 21.57 & 16.62 & 13.91 & 15.36  \\
    \textbf{GPTQT} & & \textbf{110.53} & \textbf{36.59} & \textbf{22.87} & \textbf{19.03} & \textbf{16.26} & \textbf{12.51} & \textbf{12.56} \\

    \hline
    \hline
    \end{tabular}%
  }
  \vspace{-1em}
  \label{tab:ptb}%
\end{table*}%

    Here, we primarily compare GPTQT with round-to-nearest (RTN) quantization, GPTQ, and BCQ, all of which are quantization methods for LLMs using similar technical approaches. Notably, 4-bit quantization for LLMs can be almost lossless; thus, extensive GPTQT experiments focus on 3-bit and 2-bit outcomes, as shown in Tab. \ref{tab:main}.

    In 3-bit quantization, GPTQT outperforms most of the OPT model variants from 350M to 66B. Smaller models utilizing GPTQT exhibit lower perplexity. Given that smaller models contain more compact information, this underscores GPTQT's advantages. As model size increases, the redundancy in information also increases, and the improvements from GPTQT become less pronounced. However, on opt-66B, where GPTQ shows notably poor results, GPTQT significantly reduces perplexity by 4.01.
    
    When reducing the quantization bit depth to 2-bit, both RTN and BCQ show a complete collapse in performance, whereas GPTQT still maintains reasonable perplexity, particularly in larger models. Under these stringent conditions, GPTQT substantially outperforms GPTQ, demonstrating its superior capability to extract important information during quantization. Specifically, GPTQT achieves normal perplexity with a model size of 13B, whereas GPTQ requires a model size of 66B to reach acceptable perplexity levels.

\subsection{Result on Llama2 and Bloom}
    As depicted in Tab. \ref{tab:otherModel}, GPTQT also surpasses earlier quantization methods across different types of LLMs.

    Notably, for LLMs like Llama2 that utilize GRU instead of FFN, GPTQT significantly outperforms BCQ and GPTQ. It reduces perplexity by 104.61 and 19.51 on the 7B model, and by 6e3 and 18.8 on the 13B model, respectively. It's evident that GPTQ struggles with Llama-like models, leading to severe performance deterioration at 3-bit quantization, while GPTQT effectively compensates for this deficiency. Moreover, BCQ completely fails on Llama2, highlighting GPTQT's ability to bridge the gap in binary coding methods for this model type.
    
    For Bloom models, GPTQT generally leads in performance. However, since GPTQ performs relatively well with this type of LLM, its results are not far behind those of GPTQT.

\subsection{Result on PTB dataset}

    OPT perplexity results on the PTB dataset are also provided in Tab.\ref{tab:ptb}, demonstrating that GPTQT's effectiveness is not limited to specific datasets. For the OPT model, the perplexity trends on PTB mirror those observed on WikiText2. It's important to highlight that in 3-bit quantization, the GPTQ method still experiences a notable performance decline on PTB. However, GPTQT maintains performance close to the original fp16 model on opt-30B, with only a minimal increase in perplexity of 0.72.

\subsection{Speed up of GPTQT}

    \vspace{-1em}
    \begin{table}[h]
  \centering
  \caption{The speed up of GPTQT (ms).}
  \vspace{-1em}
  \renewcommand{\arraystretch}{1.15} % increase vertical spacing
  \resizebox{\linewidth}{!}{
    \begin{tabular}{c|c|ccccc}
    \hline
    \hline
    \textbf{Method} & \textbf{Bits} & \textbf{125M} & \textbf{1.3B} & \textbf{6.7B} & \textbf{13B} & \textbf{30B} \\
    \hline
    \hline
    full& 16 & 0.69 & 1.34 & 3.03 & 5.16 & 11.0 \\
    \hline
    GPTQ & \multirow{2}[1]{*}{3}   & 1.01 & 2.01 & 2.7& 3.34 & 4.01 \\
    \textbf{GPTQT}&  & \textbf{0.76} & \textbf{1.48} & \textbf{2.00} & \textbf{2.50} & \textbf{3.21}\\

    \hline
    \hline
    \end{tabular}%
  }
  \label{tab:speed}%
\end{table}%

    In low-throughput scenarios, the performance of generative LLMs is primarily constrained by bandwidth. Weight quantization can substantially reduce communication time, which is crucial to the noticeable speed improvement seen with GPTQ. Notably, GPTQ dequantizes weights to fp16 in real-time during computations, introducing a minor computational overhead.
    
    For GPTQT, the savings in memory and bandwidth are comparable to those of GPTQ, providing it with excellent potential for efficient operation. Moreover, because its two-step quantization process can ultimately be merged into a separate binary coding, GPTQT can utilize more efficient computation methods specifically designed for binary coding, such as LUT-GEMM, resulting in more significant speed improvements as model scale increases.
    
    As detailed in Tab. \ref{tab:speed}, when the model size is small, bandwidth limitations are less apparent, and the speed of GPTQ is even considerably slower than that of fp16 models. However, as the model size increases to 30B, GPTQ enhances speed by 2.75$\times$ due to bandwidth savings. GPTQT, using the more efficient LUT-GEMM method, further boosts speed by an additional 1.24$\times$.
        
\section{Ablation Study}

In this section, we will present specific experiments to demonstrate the effectiveness of our proposed ideas and methods.

\subsection{Quantizaion Overfitting}

    \vspace{-1em}
    \begin{table}[h]
  \centering
  \caption{Overfitting on GPTQ.}
  \vspace{-1em}
  \renewcommand{\arraystretch}{1.15} % increase vertical spacing
  \resizebox{\linewidth}{!}{
    \begin{tabular}{c|cccccc}
    \hline
    \hline
    \textbf{Method} & \textbf{125M} & \textbf{350M} & \textbf{1.3B} & \textbf{2.7B} & \textbf{6.7B} & \textbf{13B} \\
    \hline
    \hline
    GPTQ(Linear Quant)  & 53.85 & 33.79 & 20.97 & 16.88 & 14.86 & 11.61 \\
    GPTQ(min MSE) &  59.38 & 35.29 & 93.76 & 73.61 & 71.53 & 15.22 \\
    GPTQ+BCQ  & 213.22 & 34.59 & 142.68 & 71.31 & 52.66 & 45.09 \\
    \textbf{GPTQT} & \textbf{90.31} & \textbf{27.41} & \textbf{17.23} & \textbf{14.88} & \textbf{14.23} & \textbf{12.98} \\

    \hline
    \hline
    \end{tabular}%
  }
  \label{tab:overfit}%
\end{table}%

    GPTQT employs a two-step quantization approach on LLMs to address the overfitting issues inherent in the GPTQ method. Specifically, within GPTQ-based strategies, the weights designated for parameter determination do not match the final quantized weights. This discrepancy necessitates a quantization method with adequate generalization capabilities.

    To illustrate this point, two specific experimental setups are introduced. \textit{GPTQ (min MSE)} utilizes linear quantization and adjusts the clipping range via grid search to minimize the MSE error between the original and quantized weights. \textit{GPTQ+BCQ} incorporates BCQ into the GPTQ framework, leveraging binary coding to decrease quantization error by creating non-uniform quantization intervals.

    Despite significantly reducing the quantization error of the weights themselves, as demonstrated in Tab.\ref{tab:overfit}, both methods perform substantially worse than both GPTQT and the original GPTQ.

\subsection{Intermediate Bit}

    In the first step, GPTQT quantizes the weight to a relatively high bit level to assess the impact of this intermediate bit depth on the results. The experiments explored quantization from 3 to 6 bits, with the second phase finalizing at 3 bits. These experiments were conducted using opt-125M, opt-350M, and opt-1.3B models, with perplexity reported on the WikiText2 dataset.

    \begin{figure}[h]
        \centering
        \includegraphics[width=0.8\linewidth]{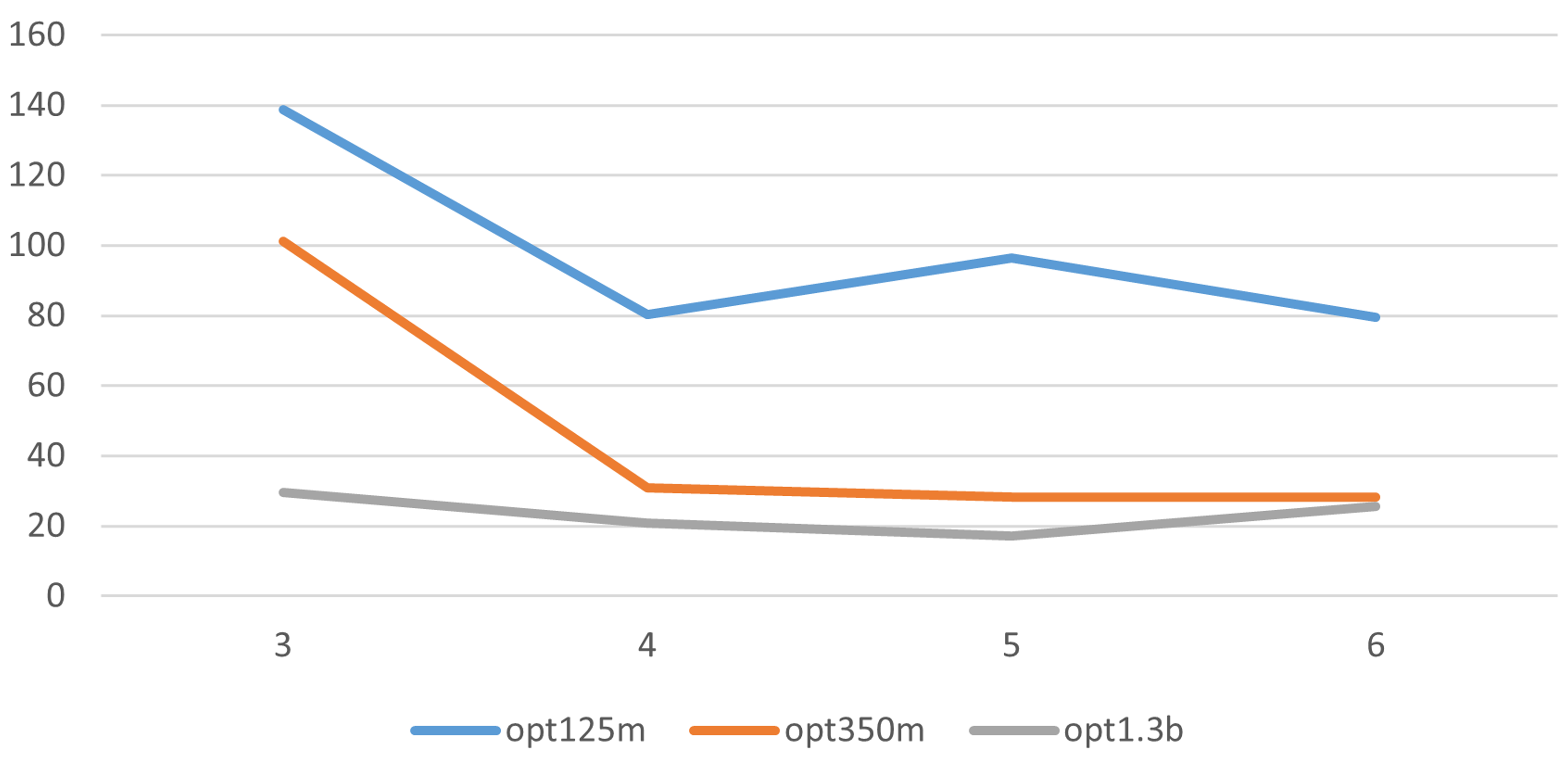}
        \caption{The impact of Intermediate Bit on results.}
        \label{fig:step1bit}
        \vspace{-0.5em}
    \end{figure}

    As indicated in Fig.\ref{fig:step1bit}, it is necessary to select a relatively high quantization bit in the first step . To strike a balance between search time and final outcome, quantizing at 4 bits or 5 bits emerges as the optimal choice.

\subsection{Re-exploration}
    \vspace{-2em}
    \begin{table}[h]
  \centering
  \caption{Result on WikiText2 with different re-exploration range.}
  \vspace{-1em}
  \renewcommand{\arraystretch}{1.15} % increase vertical spacing
  \resizebox{\linewidth}{!}{
    \begin{tabular}{c|cccc}
    \hline
    \hline
    \textbf{Range} & \textbf{opt-125M} & \textbf{opt-350M} & \textbf{opt-1.3B} & \textbf{opt-2.7B}\\
    \hline
    \hline
    0  & 399.51 & 27.41 & 17.79 & 15.08 \\
    1  &  96.41 & 28.13 & 17.23  & 14.88  \\
    2  & 90.31 & 27.51 & 20.06 & 15.91 \\
    \hline
    \hline
    \end{tabular}%
  }
  \label{tab:reexplore}%
\end{table}%

    Tab.\ref{tab:reexplore} illustrates the impact of re-exploration on the results, where "range" specifies the bit range used for exploring the scaling factor $S$. Results are reported for a configuration using a 3-bit final quantization with an intermediate 5-bit setting. Here, "Range 0" indicates no re-exploration, "Range 1" represents exploration of $S$ from 4 bits to 6 bits, and "Range 2" extends the exploration from 3 bits to 7 bits. This setup demonstrates how adjusting the range of $S$ exploration can affect the final quantization effectiveness.

\subsection{Conclusion}

In this paper, we identify overfitting as a significant issue in the quantization process with GPTQ. To address this problem, we propose a Post Train Quantization method for LLMs, named GPTQT, which divides the quantization process into two progressive steps. Initially, weights are quantized to a relatively higher bit with linear quantization. Subsequently, they are converted into a lower bit binary coding representation. To manage the change in representation range introduced by the second step, GPTQT re-explores the scaling factor to enhance performance. During inference, these two phases are merged into a pure binary coding representation, allowing the use of more efficient computational methods to achieve acceleration. However, we also note significant limitations: the activation values remain at fp16, rendering GPTQT less suitable for high-throughput applications.

\bibliography{refer}
\bibliographystyle{ieeetr}

\end{document}